\pdfoutput=1
\documentclass{article}
\usepackage{iclr2027_conference}
\iclrfinalcopy
\usepackage{times}
\usepackage[hidelinks]{hyperref}
\usepackage{url}
\usepackage{graphicx}
\usepackage{tikz}
\usetikzlibrary{arrows.meta,positioning}
\usepackage{booktabs}
\usepackage{multirow}
\usepackage{array}
\usepackage{amsmath}
\usepackage{amssymb}
\usepackage{float}
\usepackage{algorithm}
\usepackage{algpseudocode}
\usepackage{xcolor}

\usepackage{amsmath,amsfonts,bm}

\def\eqref#1{equation~\ref{#1}}

\def\1{\bm{1}}

\DeclareMathAlphabet{\mathsfit}{\encodingdefault}{\sfdefault}{m}{sl}
\SetMathAlphabet{\mathsfit}{bold}{\encodingdefault}{\sfdefault}{bx}{n}

\newcommand{\sysname}{\textsc{MEOW}}
\newcommand{\enginename}{\textsc{Lenscope}}
\newcommand{\papertitle}{Many Eyes, One World: Feed-Forward\\3D Reconstruction from Mixed Cameras}
\graphicspath{{figures/}}
\title{\papertitle}
\author{
\normalfont
Qiaoge Li$^{1}$ \quad Yifan Zhan$^{2}$ \quad Haijun Yang$^{1}$ \quad Haiyang Liu$^{2}$ \quad Yiyi Cai$^{2}$ \quad Chenchi Luo$^{1}$ \\
$^{1}$China Mobile Communications Company Limited Research Institute \\
$^{2}$The University of Tokyo
}

\begin{document}
\maketitle
\lhead{Preprint.}
\begin{figure}[H]
\centering
\includegraphics[width=\textwidth]{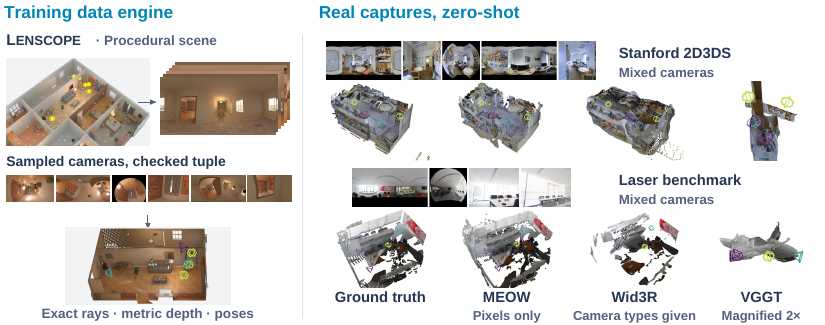}
\caption{\sysname{} reconstructs a shared 3D scene from mixed-camera images in one forward pass without supplied camera calibration or camera-type labels.
Left: \enginename{} generates mixed-camera training views from procedural scenes, checks their covisibility, and provides exact rays, metric depth, and camera poses for supervision.
Yellow markers indicate the selected panorama centres.
Right: reconstructions of mixed-camera tuples constructed from Stanford 2D3DS and laser-scanned panoramic captures.
\sysname{} receives images only, while Wid3R additionally receives camera-type labels.
Each point cloud is independently similarity-aligned to the ground truth for visualisation; VGGT panels are magnified 2$\times$.}
\label{fig:teaser}
\end{figure}

\begin{abstract}
Real-world capture is heterogeneous: perspective, fisheye, and 360$^\circ$ panoramic images can coexist within a single reconstruction task, yet most feed-forward 3D reconstruction models assume perspective imagery and a uniform input representation.
Recent models handling several camera types are either informed of the camera type for each view or reconstruct one image pair at a time.
No single-pass method reconstructs mixed-camera tuples containing full panoramas from images alone.
We present \sysname{}, a feed-forward system that jointly reconstructs metric pointmaps and camera poses from one $N$-view tuple mixing perspective, fisheye and full-panorama images, in a single forward pass from images alone: no calibration, distortion parameters, camera-type labels or poses are supplied for any view.
Our guiding design philosophy is to treat heterogeneous-camera reconstruction as a data-adaptation problem rather than an architectural redesign. 
\sysname{} retains a perspective-pretrained backbone and learns heterogeneous cameras entirely from a procedural data engine, which renders each scene across a continuous manifold of camera models with exact rays and depth, and certifies covisibility for every camera-sampled training tuple. 
Trained on synthetic tuples only, \sysname{} transfers zero-shot to real captures: on heterogeneous 2D3DS tuples it achieves 80.4 mAA@30 against 54.3 for Wid3R given the camera type of every view; on our laser-scanned mixed-camera benchmark it registers every four-view mixed tuple with 79.4 AUC@30. 
The data engine, benchmark, and complete evaluation pipeline will be released.
\end{abstract}

\section{Introduction}
\label{sec:intro}
From surveillance installations to consumer panoramas and robotic rigs, real-world capture mixes perspective, fisheye, and panoramic cameras with different projection geometries, fields of view, and aspect ratios.
Reconstructing these images together requires relating observations across camera types and recovering their shared 3D geometry, often without reliable calibration.
Feed-forward reconstruction models, including DUSt3R~\citep{dust3r}, VGGT~\citep{vggt}, $\pi^3$~\citep{pi3}, and MapAnything~\citep{mapanything}, provide a foundation for this task, but are developed primarily for perspective imagery.
Their training distributions and input processing do not adequately cover heterogeneous projections and full panoramas, limiting their performance on mixed-camera tuples (Figure~\ref{fig:teaser}, Table~\ref{tab:blk}).

Existing approaches address parts of this problem.
Rectifying wide-angle images to perspective views requires calibration and can discard field of view.
Wid3R~\citep{wid3r} jointly processes multiple camera models but requires a camera-model token for each view.
CAM3R~\citep{cam3r} removes this requirement, but processes image pairs; reconstructing a larger collection requires repeated pairwise inference followed by global alignment.
Extending joint reconstruction to mixed-camera tuples without supplied camera information presents two challenges.
First, diverse multi-view training tuples that combine perspective, fisheye, and full-panorama images with accurate geometric supervision are scarce.
Second, a shared input pipeline must accommodate their different image layouts without discarding the coverage needed for reconstruction.

We present \sysname{}, which adapts the publicly available MapAnything checkpoint to jointly reconstruct metric pointmaps and camera poses from tuples mixing perspective, fisheye, and full-panorama images.
Reconstruction takes one network forward pass, without externally supplied calibration, distortion parameters, camera-type labels, or poses.
Our approach retains the perspective-pretrained backbone and uses synthetic data for all subsequent adaptation.
\enginename{} supplies geometrically supervised mixed-camera training tuples, while lightweight input adaptations preserve the coverage and layout information needed to process these views together.

\enginename{} generates camera diversity from procedurally constructed indoor scenes (Section~\ref{sec:engine}).
At each sampled optical centre, we render a full panorama with exact rays, metric depth, and camera pose, then resample it into perspective, fisheye, or panoramic views with varying camera parameters.
This construction varies field of view, distortion, and image layout while retaining geometric supervision.
Camera sampling also changes which parts of the scene remain visible: two overlapping panoramic observations may produce narrow views with little shared content.
We therefore recompute covisibility after sampling the cameras and retain tuples whose views form a connected covisibility graph.
The resulting training stream combines diverse camera projections with overlapping observations that support joint reconstruction.

To process heterogeneous views together, we resize every image to a common tensor shape without cropping (Section~\ref{sec:contract}).
This preserves the complete image coverage, including all longitudes of a full panorama.
Resizing can change aspect ratios, however, so the resulting tensor shape no longer identifies the source layout.
An aspect-ratio embedding supplies this information to the network, while circular padding in the dense prediction head handles the longitude boundary of full panoramas.
These panoramas are identified from their images at inference, without supplied camera-type labels.
We also adapt geometric supervision to the different projections: solid-angle weighting accounts for unequal directional coverage per pixel, and a directional likelihood supervises predicted rays (Section~\ref{sec:recipe}).

We evaluate \sysname{} on mixed-camera tuples constructed from real panoramic captures, with perspective and fisheye views resampled from the original panoramas.
On heterogeneous Stanford 2D3DS tuples, \sysname{} reaches 80.4 mAA@30, compared with 54.3 for Wid3R supplied with camera-type labels.
On our laser-scanned benchmark, where all methods are evaluated without fine-tuning, \sysname{} reaches 79.4 AUC@30 on mixed tuples, compared with 29.3 for Wid3R.
Controlled comparisons show that synthetic adaptation substantially improves the pretrained backbone, whereas changing its input resizing alone does not.
Further ablations show that full-field-of-view resizing and the aspect-ratio embedding contribute to the adapted model's performance.

Our contributions are threefold:
\begin{itemize}
\item We adapt a perspective-pretrained reconstruction model to jointly recover metric pointmaps and camera poses from mixed-camera tuples containing full panoramas, in one forward pass and without externally supplied calibration, distortion parameters, camera-type labels, or poses.
\item We develop \enginename{}, a reproducible procedural data engine that generates heterogeneous camera views with exact geometric supervision and checks covisibility after camera sampling, enabling adaptation using synthetic tuples only.
\item We introduce a laser-scanned benchmark with mixed-camera and single-camera tracks, and evaluate reconstruction methods without fine-tuning on it.
We will release the data engine, benchmark, and evaluation pipeline.
\end{itemize}

\section{Related Work}
\label{sec:related}
\paragraph{Feed-forward reconstruction.}
Feed-forward reconstruction has moved from image pairs to large view collections.
DUSt3R~\citep{dust3r} predicts pointmaps from two images, and MASt3R~\citep{mast3r} adds matching.
MUSt3R, Fast3R, and CUT3R~\citep{must3r,fast3r,cut3r} extend this line to many views, while VGGT and $\pi^3$~\citep{vggt,pi3} predict joint $N$-view geometry.
MapAnything~\citep{mapanything} adds metric scale and optional geometric cues; Depth Anything 3~\citep{depthanything3} accommodates variable view counts.
Yet these models remain centred on perspective images in a common tensor layout.
Rig3R~\citep{rig3r} uses rig metadata when available and infers it otherwise, while Pow3R~\citep{pow3r} accepts optional camera and depth priors.
We build on the MapAnything backbone.

\paragraph{Non-pinhole cameras in feed-forward reconstruction.}
Existing methods differ chiefly in what camera information they require.
Wid3R~\citep{wid3r} needs a per-view camera token, which UCE~\citep{uce} infers from the image.
Rig3R~\citep{rig3r} can infer missing rig structure but targets perspective rigs; Fisheye3R~\citep{fisheye3r} gates calibration tokens by camera type; and X-Lens~\citep{xlens} takes per-pixel calibration and a camera type for every view and predicts metric depth but no poses.
CAM3R~\citep{cam3r} needs no calibration, but reconstructs two views at a time and aligns them afterward.
PanoVGGT, CasaMaestro, and Argus~\citep{panovggt,casamaestro,argus} handle panoramas alone, while RIGOR and HALO-SLAM~\citep{rigor,haloslam} use frozen foundation models for panoramic SLAM.
Ray-aware alternatives use supplied rays for mixed pinhole/fisheye views~\citep{gray}, adapt to one fisheye camera per sequence~\citep{raytun3r}, or address novel-view synthesis and calibrated rig depth~\citep{rore,pfdepth}.
Among the systems in Table~\ref{tab:contracts}, \sysname{} alone combines mixed tuples containing full panoramas, no supplied calibration, distortion, or camera-type labels, one joint forward pass, and metric pointmaps with poses.
\begin{table}[b]
\centering
\caption{Input requirements and outputs of the closest systems, read from their papers and code (all cited in Section~\ref{sec:related}). \checkmark: supported; (\checkmark): partial or qualitative by the authors' own account; --: not supported or not claimed. Only \sysname{} takes a mixed tuple with full panoramas from pixels alone and returns metric pointmaps and poses in one pass.}
\label{tab:contracts}
\footnotesize
\setlength{\tabcolsep}{3pt}
\renewcommand{\arraystretch}{1.0}
\begin{tabular}{@{}llccccc@{}}
\toprule
Method & Camera input at test time & Mixed tuple & Full 360$^\circ$ & $N$-view pass & Poses & Metric \\
\midrule
Wid3R & Class per view & \checkmark & \checkmark & \checkmark & \checkmark & -- \\
CAM3R & None & (\checkmark) pairs & \checkmark & -- & \checkmark & -- \\
Fisheye3R & None & (\checkmark) & (\checkmark) & \checkmark & \checkmark & -- \\
X-Lens & Intrinsics or rays, and class & \checkmark & -- & \checkmark & -- & \checkmark \\
G-ray & Ray map per view & \checkmark & -- & \checkmark & \checkmark & -- \\
PanoVGGT & Panoramas only & -- & \checkmark & \checkmark & \checkmark & -- \\
\sysname{} (ours) & None & \checkmark & \checkmark & \checkmark & \checkmark & \checkmark \\
\bottomrule
\end{tabular}
\end{table}

\paragraph{Monocular any-camera geometry and calibration.}
Monocular depth has expanded beyond perspective cameras~\citep{unik3d,dac,unidac,da2,dap,calibtokens}; OmniPoint~\citep{omnipoint} and PaGeR~\citep{pager} recover metric geometry from an arbitrary-camera image and a panorama, respectively.
Another line estimates the camera model from one image~\citep{geocalib,anycalib,deepcalib,babelcalib} or from sparse views~\citep{calibanyview}. 
We let multiple views constrain one another, jointly recovering geometry and poses without supplied camera parameters.

\paragraph{Camera models and synthetic data.}
Our camera manifold draws on classical models~\citep{kannala,geyer,eucm,doublesphere,mei}.
Synthesising cameras from panoramas~\citep{deepcalib,cam3r,wid3r} and generating procedural multi-view data~\citep{simpleproc} are established practices.
Existing resources span fixed indoor scenes (Structured3D, PanoCity, OmniRooms~\citep{structured3d,panovggt,unisharp}), procedural scenes (PanoInfinigen~\citep{pager}), curated asset views (CM-EVS~\citep{cmevs}), and real sequences with laser geometry (Holo360D~\citep{holo360d}).
We generate scenes reproducibly from seeds using procedural geometry and CC0 textures on a single GPU.
Our engine samples a continuous camera manifold with photographic priors over principal point, roll, tilt, and optical blur, checks covisibility for each tuple, and outputs exact rays and depth for views with distinct optical centres.

\section{Method}
\label{sec:method}

\subsection{Problem Formulation}
\label{sec:problem}
Given $N$ images captured by unknown and potentially different camera models, our goal is to reconstruct metric pointmaps and camera poses in a single forward pass.
Writing $I_i\in\mathbb{R}^{H_i\times W_i\times 3}$ for the $i$-th image, \sysname{} is one network $f_\theta$ with
\begin{equation}
f_\theta\big(\{I_i\}_{i=1}^{N}\big)=\Big(\{\mathbf{r}_i,\,d_i,\,\mathbf{R}_i,\,\mathbf{t}_i\}_{i=1}^{N},\ s\Big),\qquad
\mathbf{X}_i(\mathbf{p})=s\,\big(\mathbf{R}_i\,d_i(\mathbf{p})\,\mathbf{r}_i(\mathbf{p})+\mathbf{t}_i\big),
\label{eq:task}
\end{equation}
where $\mathbf{r}_i(\mathbf{p})\in\mathbb{S}^2$ is the unit ray of pixel $\mathbf{p}$ in the camera frame of view $i$, $d_i(\mathbf{p})>0$ the depth along that ray, $(\mathbf{R}_i,\mathbf{t}_i)$ the camera-to-world pose in the frame of the first view (OpenCV convention), $s>0$ one metric scale for the tuple and $\mathbf{X}_i$ the resulting metric pointmap.
At inference, no camera types, intrinsics, distortion parameters, or poses are externally supplied; the only auxiliary cues are each image's native shape $(H_i,W_i)$ and a binary panorama flag $\pi_i$ inferred from its pixels.
We study whether high-quality synthetic data can lift a perspective-pretrained geometry model to this mixed-camera task.
Starting from the publicly available MapAnything checkpoint, we perform all subsequent adaptation on synthetic mixed-camera tuples.
We first present \enginename{}, our synthetic data engine (Section~\ref{sec:engine}); then describe how images retain their full field of view within a shared tensor shape (Section~\ref{sec:contract}); and finally detail the model, losses and training procedure (Section~\ref{sec:recipe}).

\subsection{\enginename{}: the camera-manifold engine}
\label{sec:engine}
\begin{figure}[t]
\centering
\includegraphics[width=\textwidth]{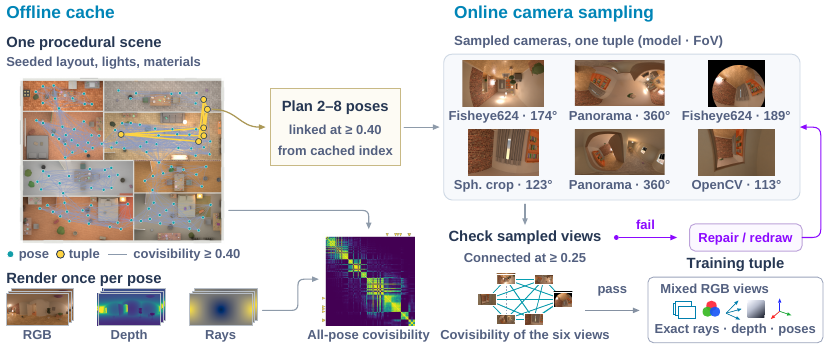}
\caption{Offline, once per scene: a seed generates a procedural scene, panorama poses are placed in its free space, each pose is rendered with exact rays and depth; pairwise covisibility is computed by mutual visibility tests (lines: $c_{kl}\ge0.40$).
Online, per tuple: a connected walk plans 2--8 poses and each is resampled through a sampled camera.
Covisibility is recomputed (edge width; dashed below 0.25) and only connected tuples are kept.}
\label{fig:enginediag}
\end{figure}
Figure~\ref{fig:enginediag} draws the pipeline: an offline stage renders and indexes every scene once, and an online stage assembles one tuple per training sample.
Appendix~\ref{app:engine} lists the parameters and gives the complete procedure as Algorithm~\ref{alg:delivery}.
Scenes come from our own procedural indoor generator, in the spirit of ProcTHOR~\citep{procthor}, in two generations: 2,004 first-generation rooms and 499 denser and more realistic second-generation scenes.
Panorama poses are placed in the free space of each scene, and every pose $k$ is rendered once as an equirectangular panorama $E_k$ with analytic unit rays, radial depth and a validity mask.
Panoramic sources allow fields of view up to $360^\circ$ with exact rays and depth.
Fisheye views that are warped from perspective frames, by contrast, cannot exceed the field of view of their source.

\paragraph{Covisibility.}
Two forms of one visibility test are used.
Offline, between the panorama poses $k,l$ of a second-generation scene, $\mathrm{cov}(k\!\to\!l)$ is estimated with 96 directions drawn uniformly on the sphere at pose $k$.
Each direction is cast to the nearest solid, unambiguous surface point, which counts as covisible if the straight line from pose $l$ reaches it unoccluded (within 5\,cm).
The stored value is $c_{kl}=\min\big(\mathrm{cov}(k\!\to\!l),\mathrm{cov}(l\!\to\!k)\big)$, an equal-solid-angle estimate that does not depend on the panorama grid.
Online, on the sampled views, the test runs on the pixel grid of each view rendered at 96 pixels on the long side.
A valid pixel $\mathbf{p}$ of view $i$ is covisible in $j$ when its 3D point, expressed in the frame of $j$ as $\mathbf{Y}=\mathbf{R}_j^{\!\top}\big(\mathbf{X}_i(\mathbf{p})-\mathbf{t}_j\big)$, lies in front of $j$, falls within $3.6^\circ$ of the nearest ray $\mathbf{q}^\star=\arg\max_{\mathbf{q}}\langle\mathbf{Y}/\|\mathbf{Y}\|,\mathbf{r}_j(\mathbf{q})\rangle$ of a valid pixel, and is not occluded there:
\begin{equation}
\big|\,\|\mathbf{Y}\|-d_j(\mathbf{q}^\star)\big|\le 0.10\,\mathrm{m}+0.05\,d_j(\mathbf{q}^\star).
\label{eq:covis}
\end{equation}
$\mathrm{cov}(i\!\to\!j)$ is the fraction of valid pixels of $i$ that pass, and the online check uses $c_{ij}=\tfrac12\big(\mathrm{cov}(i\!\to\!j)+\mathrm{cov}(j\!\to\!i)\big)$.
The two directions differ by construction: a narrow view in the forward hemisphere of a panorama is covered entirely while the panorama is not.
Each direction is normalised by its own sample count, so views of different resolution and field of view are comparable.
Covisibility is recomputed online because a narrow or tilted camera sees a fraction of what its panorama saw.

\paragraph{Tuple sampling and camera manifold.}
A tuple of $K\in[2,8]$ poses with distinct optical centres is drawn by a random walk on the offline matrix that adds a pose only if its covisibility with the current pose is at least 0.40 (relaxed to 0.20 in the aimed mode below when no chain of length $K$ exists).
With probability 0.55 a tuple is planned in an aimed mode, in which the cameras converge on a common surface point (probability 0.90) or look along a shared direction toward it; otherwise each camera keeps the orientation of its panorama pose. A plan that fails is retried in the other mode.
Each pose is then resampled through a camera drawn from seven models with coupled field-of-view ranges (Table~\ref{tab:cammodels}, Appendix~\ref{app:engine}).
Rectilinear OpenCV and pinhole cameras form the majority; Fisheye624, EUCM and Mei cover the fisheye family; spherical crops and full panoramas cover the sphere.
Initial fields of view are log-normal around an 80$^\circ$ diagonal for the rectilinear models and log-uniform over the model's range for the others, and distortion coefficients are drawn from ranges valid for that field of view.
Full panoramas receive a random $SO(3)$ content rotation.
Principal-point shifts and roll follow priors set from calibration statistics and photographic collections~\citep{clarke1998,holdgeoffroy2023}.
An optics layer (modulation transfer, vignetting, sensor noise) is decoupled from geometry, so blur is a weaker cue to the field of view.

\paragraph{Covisibility check and repair.}
The sampled views are rendered at low resolution, $c_{ij}$ is recomputed with Eq.~(\ref{eq:covis}), and the tuple is kept only if the graph with edges $c_{ij}>0.25$ is connected.
A disconnected view is moved in three steps toward a wide field of view, a square aspect and no tilt or principal-point offset, and the graph is re-tested; a tuple that still fails is discarded.
Kept views are rendered once more at their target resolution.
No real image of any camera type enters training after the public perspective pretraining.
Scene realism matters: under the same configuration, second-generation scenes instead of first-generation renders lower Matterport3D~\citep{matterport3d} zero-shot accuracy and completeness error by 17.4\% and 24.1\%.
Table~\ref{tab:datacmp} (Appendix~\ref{app:tables}) contrasts this construction with the training data of the compared methods, read from their released code.

\subsection{Mixed-camera inputs}
\label{sec:contract}
\paragraph{Resizing without cropping.}
All views of a tuple enter the network at one tensor shape $(h,w)$ drawn from ten buckets $\mathcal{B}$ whose sides are multiples of 14 and whose width-to-height ratios range from 0.49 to 3.08.
View $i$ is resized anisotropically, $\tilde I_i(u,v)=I_i\!\left(uW_i/w,\;vH_i/h\right)$, so that the tensor grid covers the entire native image and nothing is cropped (Figure~\ref{fig:contract}, Appendix~\ref{app:tables}).
Ground-truth rays, depth and masks are resampled through the same map, and rays keep their directions.
During training the bucket is drawn per batch independently of the content, so the deformation $\delta_i=(w/h)\,/\,(W_i/H_i)$ of each view is random.
At inference the bucket closest to the mean aspect ratio of the tuple is used for all views.

\paragraph{Aspect-ratio embedding.}
Since the bucket is drawn independently of the content, the tensor shape no longer tells the network the native aspect of a view.
A small MLP $g$ with a zero-initialised last layer maps $a_i=\log(W_i/H_i)$ to a vector that is added to every patch token of view $i$, $\mathbf{F}_i\leftarrow\mathbf{F}_i+\mathbf{1}\,g(a_i)^{\!\top}$, following the global-representation encoder of MapAnything.
The embedding holds 0.08\% of the parameters.

\paragraph{Panorama flag, wrap and detection.}
A full equirectangular panorama is periodic in longitude.
For views with $\pi_i=1$ the dense head pads its token grid with three circular columns on each side, runs, and crops back, so no parameter is added, and the aspect input takes its definitional value $a_i=\log 2$ whatever the shape in which the panorama arrives; other views run the head unchanged.
In training $\pi_i$ is read from the ground-truth rays (azimuth span of the middle row above $350^\circ$); at inference it comes from a detector.\label{sec:detector} A full equirectangular image has two signatures that survive any aspect squeeze: its last and first columns are adjacent in content, and its top and bottom rows compress all longitudes into near-constant colour.
The detector measures both on a 128-row copy of the image as ratios against its interior, with a guard against circular fisheyes with black corners; its thresholds were fixed on training renders only (Appendix~\ref{app:detector}).
On the two headline benchmarks it misses no panorama and flags 2 of 386 other views (Table~\ref{tab:detector}, Appendix~\ref{app:detector}).

\subsection{Model, training and inference}
\label{sec:recipe}
\begin{figure}[t]
\centering
\includegraphics[width=\textwidth]{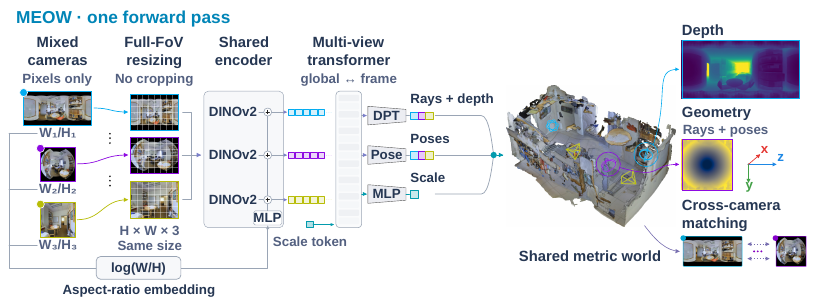}
\caption{One forward pass.
Views from different cameras are resized to one tensor shape with their full field of view; the native aspect of each (fixed to 2:1 for detected full panoramas) enters through an MLP added to its patch tokens.
Sixteen alternating global and frame attention layers, a dense head per view (circular padding for panoramas), a pose head and a scale head give rays, depth, camera-to-world poses (OpenCV axes) and metric scale in one shared world.
Inputs: a real tuple, zero-shot.}
\label{fig:overview}
\end{figure}
\paragraph{Architecture.}
Figure~\ref{fig:overview} presents the overall system, whose reconstruction network retains the original MapAnything backbone without architectural
modification.
The first 24 layers of a DINOv2~\citep{dinov2} ViT-G encoder, shared across views, turn each resized view into patch tokens $\mathbf{F}_i\in\mathbb{R}^{(hw/14^2)\times1536}$, to which $g(a_i)$ is added.
16 transformer layers (width 1536, 24 heads) alternate global attention over the tokens of all views together with one learnable scale token, and frame attention within each view.
One DPT~\citep{dpt} dense head, applied to every view with the wrap of Section~\ref{sec:contract}, predicts $\mathbf{r}_i$, $d_i$, a confidence and a validity logit.
One pose head per view (convolutional pooling and an MLP) predicts $\mathbf{R}_i$ as a unit quaternion and $\mathbf{t}_i$, and a scale head on the scale token predicts $s$.
Pointmaps follow from Eq.~(\ref{eq:task}).

\paragraph{Losses.}
Ground truth carries an asterisk and $m_i$ is the valid mask of view $i$.
Two changes make the MapAnything objective aware of non-perspective geometry.
First, the ray, depth and point terms of view $i$ are weighted by the solid angle of its ground-truth ray field,
\begin{equation}
\omega_i(\mathbf{p})=\frac{\big\|\partial_u\mathbf{r}^*_i\times\partial_v\mathbf{r}^*_i\big\|(\mathbf{p})}{\operatorname{mean}_{\mathbf{p}'}\big\|\partial_u\mathbf{r}^*_i\times\partial_v\mathbf{r}^*_i\big\|(\mathbf{p}')},\qquad \omega_i\in[0.05,\,20],
\label{eq:solid}
\end{equation}
normalised to mean one over all pixels of each view, so that the poles of a panorama and the rim of a fisheye, which the pixel grid over-represents, no longer dominate; on a pinhole view $\omega_i$ is proportional to $\cos^3$ of the off-axis angle.
Second, ray directions are scored by a von Mises--Fisher likelihood on $\mathbb{S}^2$~\citep{mardia} with one fixed concentration $\kappa=e^3$,
\begin{equation}
\mathcal{L}_{\mathrm{ray}}=\sum_i\sum_{\mathbf{p}}\omega_i(\mathbf{p})\Big[-\kappa\,\big\langle\mathbf{r}_i(\mathbf{p}),\mathbf{r}^*_i(\mathbf{p})\big\rangle+\log\frac{4\pi\sinh\kappa}{\kappa}\Big],
\label{eq:vmf}
\end{equation}
which, with $\kappa$ fixed, is a cosine loss scaled by $\kappa$ plus a constant.
The remaining terms follow MapAnything.
With $\rho(\mathbf{e})=\frac{|\alpha-2|}{\alpha}\big[\big(\frac{\|\mathbf{e}\|^2/c^2}{|\alpha-2|}+1\big)^{\alpha/2}-1\big]$ the robust penalty of~\citet{barron} ($\alpha=0.5$, $c=0.05$), applied in log space after predictions and targets are normalised by their mean distance to the origin,
\begin{equation}
\begin{aligned}
\mathcal{L}=\sum_i\Big[\rho\big(\mathbf{X}_i-\mathbf{X}^*_i\big)&+0.1\,\big(\rho(\mathbf{P}_i-\mathbf{P}^*_i)+\rho(d_i-d^*_i)+\rho(\mathbf{q}_i-\mathbf{q}^*_i)+\rho(\mathbf{t}_i-\mathbf{t}^*_i)\big)\\
&+0.3\,\big(\mathcal{L}_{\mathrm{normal}}+\mathcal{L}_{\mathrm{grad}}\big)\Big]+0.1\,\rho(s-s^*)+0.1\,\mathcal{L}_{\mathrm{ray}}+0.03\,\mathcal{L}_{\mathrm{mask}},
\end{aligned}
\label{eq:loss}
\end{equation}
where dense terms are averaged over the valid pixels of each view position across the tuples of a batch (the ray term over all pixels) and the loss is scaled by $2/K$, $\mathbf{P}_i=d_i\mathbf{r}_i$ is the camera-frame pointmap, the pose terms are applied to the absolute poses and to all pairwise relative poses (the quaternion term to the nearer of $\mathbf{q}^*_i$ and $-\mathbf{q}^*_i$), $\mathcal{L}_{\mathrm{normal}}$ compares normals of the camera-frame pointmaps, $\mathcal{L}_{\mathrm{grad}}$ matches log-depth gradients, and $\mathcal{L}_{\mathrm{mask}}$ is a binary cross-entropy on the validity logit.
The world-point term is confidence-weighted as in DUSt3R ($c_{\mathbf{p}}\rho-0.2\log c_{\mathbf{p}}$).

\paragraph{Training and inference.}
Training starts from the public MapAnything checkpoint.
Stage 1 fine-tunes the checkpoint on initial synthetic renders with the official pipeline for 35 epochs.
Stage 2 turns on the full-field-of-view resizing, the aspect-ratio embedding and the online camera sampling of Section~\ref{sec:engine} (100 epochs).
Stage 3 trains on second-generation scenes with the losses of Eqs.~(\ref{eq:solid})--(\ref{eq:loss}) and the panorama wrap, in two runs from Stage 2: one of 15 epochs and one of 859 epochs; the final weights interpolate the two (Appendix~\ref{app:training}).
Batches hold tuples of 2--8 views at the ten aspect buckets with up to 48 images per GPU on four H200 GPUs (at most 25 on eight); an epoch has 416 steps per GPU and takes 33 minutes on four H200s in Stage 2.
For inference, the $N$ images are resized to the bucket closest to their mean aspect ratio, $a_i$ is read from their native shapes (set to $\log 2$ for detected full panoramas) and $\pi_i$ from the detector, and one forward pass returns Eq.~(\ref{eq:task}); no alignment, matching or optimisation follows.

\section{Experiments}
\label{sec:exp}
\begin{figure}[t]
\centering
\includegraphics[width=0.92\textwidth]{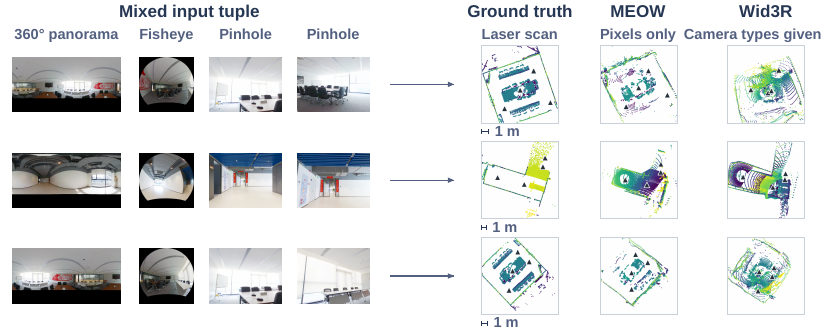}
\caption{Laser-scanned mixed tuples, zero-shot.
Each row shows the four inputs, the laser ground truth and the fused pointmaps of \sysname{} (pixels only) and Wid3R (camera types given), placed by the scorer's alignment.
Top-down floor-to-wall slices coloured by height; triangles mark stations or predicted cameras.}
\label{fig:blkqual}
\end{figure}

\subsection{Protocols}
\label{sec:protocols}
Both real benchmarks start from real panoramic captures: the panoramas are the originals, and the perspective and fisheye views are resampled from them through the stated camera models, so every view carries exact rays and metric ground truth.
Pose metrics are computed over ordered view pairs up to a 30$^\circ$ threshold: relative rotation and translation accuracy (RRA@30, RTA@30; sign-agnostic translation direction,~\citealp{posediffusion}), their mean average accuracy (mAA@30) and area under the curve (AUC@30); ATE is the absolute trajectory error after one Sim(3) alignment per tuple (definitions in Appendix~\ref{app:protocols}).
Every baseline runs with its official loader and checkpoint on the same tuples, ground truth and scorer; pointmap metrics and further protocol details are in Appendix~\ref{app:protocols}.
\subsection{Heterogeneous tuples}
\label{sec:t1}
Table~\ref{tab:t1} reports the heterogeneous 2D3DS benchmark: areas 5a, 5b and 6 of Stanford 2D3DS~\citep{s2d3ds}, 88 tuples of 3--24 views cycling through full panoramas, synthesised $90^\circ$ perspective views and $180^\circ$ equidistant fisheyes, all views in one forward pass.
\sysname{} reaches 80.4 mAA@30; Wid3R, given the camera class of every view, reaches 54.3, and the difference lies in translation (RTA@30 96.2 against 86.5, ATE 0.62 against 0.83).
The perspective models, which also receive pixels only, reach 10.5--19.8.
Input resizing alone does not explain the gain: the public MapAnything weights score 16.4 with their own loader and 15.4 with our full-field-of-view resizing (Table~\ref{tab:controls}, Appendix~\ref{app:tables}).

\begin{table}[b]
\centering
\caption{Heterogeneous 2D3DS tuples \citep{s2d3ds}: 88 tuples of 3--24 views, one forward pass. Pose metrics are computed per tuple and averaged over the 88 tuples; ATE is after Sim(3) alignment. $^\dagger$No multi-view weights released.}
\label{tab:t1}
\footnotesize
\setlength{\tabcolsep}{6pt}
\begin{tabular}{@{}llcccc@{}}
\toprule
Method & Camera input & RRA@30 & RTA@30 & mAA@30 & ATE $\downarrow$ \\
\midrule
VGGT & None & 32.6 & 47.3 & 10.5 & 1.65 \\
$\pi^3$ & None & 45.9 & 59.7 & 19.8 & 1.24 \\
MapAnything & None & 52.8 & 53.9 & 16.4 & 1.48 \\
CAM3R$^\dagger$ & None & -- & -- & -- & -- \\
\sysname{} (ours) & None & 95.2 & \textbf{96.2} & \textbf{80.4} & \textbf{0.62} \\
\midrule
Wid3R & Class per view & \textbf{96.9} & 86.5 & 54.3 & 0.83 \\
\bottomrule
\end{tabular}
\end{table}

\subsection{A laser-scanned benchmark on which every method is zero-shot}
\label{sec:blk}
We will release a four-track benchmark built from our own registered BLK360 G2 laser scans of a multi-room office: 12 stations with survey-grade poses and pointmaps, tuples selected by covisibility computed from the scans, and regions the scanner cannot see masked out; no method has trained on it.
Table~\ref{tab:blk} gives the results and Figure~\ref{fig:blkqual} shows three mixed tuples.
On the four-view mixed track \sysname{} registers every tuple (RRA@30 and RTA@30 of 100) and reaches 79.4 AUC@30 against 19.0--32.9 for the open baselines, which collapse on the panorama track.
Wid3R registers the rotations of the mixed tuples but not their translations (RTA@30 58.0, AUC@30 29.3), and its pointmaps are less complete.

\begin{table}[t]
\centering
\caption{Laser-scanned benchmark, 24 four-view tuples per track. AUC@30 is the mean over tuples; Acc and Comp are in metres after one alignment per tuple; NC is normal consistency.}
\label{tab:blk}
\scriptsize\renewcommand{\arraystretch}{0.9}
\setlength{\tabcolsep}{3pt}
\begin{tabular}{@{}llcccccc ccc@{}}
\toprule
 & & \multicolumn{6}{c}{Mixed tuples (panorama + fisheye + two pinholes)} & \multicolumn{3}{c}{Single-camera tuples, AUC@30} \\
\cmidrule(lr){3-8}\cmidrule(lr){9-11}
Method & Camera input & RRA@30 & RTA@30 & AUC@30 & Acc $\downarrow$ & Comp $\downarrow$ & NC & Panorama & Pinhole & Fisheye \\
\midrule
DUSt3R & None & 62.5 & 67.4 & 32.9 & 0.164 & 0.863 & 0.758 & 0.5 & 96.0 & 73.8 \\
MASt3R & None & 66.7 & 66.0 & 31.8 & 0.182 & 0.508 & 0.737 & 2.7 & 96.2 & 56.2 \\
VGGT & None & 45.8 & 63.9 & 19.0 & 0.279 & 0.591 & 0.665 & 1.9 & 97.1 & 55.0 \\
$\pi^3$ & None & 66.0 & 63.9 & 26.6 & 0.240 & 0.693 & 0.736 & 1.6 & \textbf{97.9} & 75.1 \\
MapAnything & None & 66.0 & 57.3 & 24.5 & 0.255 & 0.830 & 0.653 & 1.5 & 86.8 & 75.1 \\
\sysname{} (ours) & None & \textbf{100} & \textbf{100} & \textbf{79.4} & \textbf{0.142} & \textbf{0.367} & \textbf{0.809} & 71.0 & 85.7 & 73.2 \\
\midrule
Wid3R & Class per view & 97.2 & 58.0 & 29.3 & 0.199 & 0.467 & 0.764 & 81.9 & 90.5 & \textbf{87.3} \\
PanoVGGT & Panoramas & -- & -- & -- & -- & -- & -- & \textbf{90.0} & -- & -- \\
\bottomrule
\end{tabular}
\end{table}

\subsection{Single-camera inputs and efficiency}
\label{sec:pano}\label{sec:persp}\label{sec:efficiency}
On single-camera input the model behaves like its backbone on pinholes (85.7 against 86.8 AUC@30 on the laser pinhole track, Table~\ref{tab:blk}) and, unlike any perspective model, registers panorama-only tuples: 75.4 AUC@30 on 2D3DS and 71.0 on the laser track against at most 8.4 (Table~\ref{tab:realpano}, Appendix~\ref{app:tables}).
Inference costs what MapAnything costs on the same views, 0.66\,s for eight views on one RTX 5090 with image loading included, and 32 views fit in one network forward pass (Table~\ref{tab:efficiency}, Appendix~\ref{app:tables}).

\subsection{Ablations}
\label{sec:ablations}\label{sec:recipecmp}
Table~\ref{tab:ablation} changes one thing at a time.
On the final weights, the public crop loader in place of the full-field-of-view resizing costs 3.3 mAA@30 on 2D3DS and 5.0 AUC@30 on the laser mixed track, and withholding the aspect-ratio embedding costs 23.2 and 11.3, with the largest drop on 16:9-squeezed panoramas, where the embedding carries the content aspect.
Stage 2, where the embedding enters, was trained in two runs that differ only in it (Table~\ref{tab:phaseb}, Appendix~\ref{app:tables}): the run with the embedding leads by 3.2 mAA@30, 3.7 AUC@30 and 9.9 AUC@30 on 16:9 panoramas, trails by 1.7 on the laser panorama track, and is the one carried into Stage 3.

\vspace{-6pt}
\begin{table}[H]
\centering
\caption{Ablations. Top: the final weights with one input change at a time; the crop-loader row routes panoramas by the dataset annotation, with which the full model scores as with the detector (Table~\ref{tab:controls}). Bottom: the two Stage-2 runs, identical except for the aspect-ratio embedding (Table~\ref{tab:phaseb}). Columns: per-tuple mAA@30 on the 88 heterogeneous 2D3DS tuples; laser mixed-track AUC@30; AUC@30 on 2D3DS panorama tuples squeezed to 16:9.}
\label{tab:ablation}
\scriptsize\renewcommand{\arraystretch}{0.9}
\setlength{\tabcolsep}{4pt}
\begin{tabular}{@{}p{6.4cm}ccc@{}}
\toprule
 & 2D3DS mAA@30 & Laser mixed AUC@30 & 16:9 panoramas AUC@30 \\
\midrule
Final weights (full model) & 80.4 & 79.4 & 78.6 \\
\quad Crop loader instead of full-field-of-view resizing & 77.1 & 74.4 & -- \\
\quad Aspect-ratio embedding withheld & 57.2 & 68.1 & 42.4 \\
\midrule
Stage-2 run with the embedding (carried into Stage 3) & 63.8 & 65.4 & 63.8 \\
Stage-2 run without the embedding & 60.6 & 61.7 & 53.9 \\
\bottomrule
\end{tabular}
\end{table}

\vspace{-6pt}

\section{Conclusion}
\label{sec:conclusion}\label{sec:analysis}\label{sec:corrmain}\label{sec:limitations}
We presented \sysname{}, a feed-forward model that reconstructs metric pointmaps and camera poses from one tuple of perspective, fisheye and full-panorama images in a single forward pass, with no camera information supplied for any view.
A perspective-pretrained backbone is adapted, with full-field-of-view resizing, on \enginename{} tuples rendered through a continuous manifold of camera models with exact rays and depth and checked for covisibility.
Adapted using synthetic tuples only, the model transfers zero-shot: 80.4 mAA@30 on heterogeneous 2D3DS tuples against 54.3 for Wid3R given the camera type of every view, and every tuple on the four-view mixed track of our laser-scanned benchmark registered at 79.4 AUC@30.
The ablations show what it relies on: the full-field-of-view resizing and the aspect-ratio embedding at inference, and the embedding already at the stage where it enters training.
The shared metric world also yields cross-camera correspondences without a matcher (Appendix~\ref{app:corr}).
\paragraph{Limitations.}
The engine covers synthetic indoor scenes, and the final model is adapted on room-scale tuples of up to eight views: on 2D3DS tuples of 15--24 views it trails Wid3R, which puts Wid3R 0.5 mAA@30 ahead when pairs are pooled over all tuples (Appendix~\ref{app:protocols}), and on centimetre-baseline perspective video (Replica~\citep{replica}, ADT~\citep{adt}) it trails the perspective models (Table~\ref{tab:realpersp}, Appendix~\ref{app:tables}).
The scale-invariant metrics do not test the predicted metric scale, which is 12--25\% short of the truth (Appendix~\ref{app:protocols}).
Extending the training range to longer sequences and replaying real perspective video during adaptation are the next steps.

\newpage  %
\subsubsection*{AI use statement}
The research ideas, the experimental design, the claims and the story of this paper are the authors'. Generative AI tools assisted with the manuscript and citation checks, \LaTeX{} editing, schematic artwork, and benchmarking code. The authors verified the code, re-derived every reported number from the raw logs, checked the citations, reviewed the final manuscript, and take responsibility for the final content. No generative model was used to produce training data.

\subsubsection*{Reproducibility statement}
All randomness in the engine, the training runs and the benchmark construction is seeded (the sampler's feasibility tables also depend on Python's unrecorded string-hash seed; the tables used in training will be released), and both scene generations regenerate from seeds with scripts that will be released with the code: a regeneration of three first-generation scenes reproduced the rays, depth and validity masks of the original training packs exactly, with RGB differing only by renderer sampling noise (PSNR above 75\,dB).
The engine, the benchmark construction scripts with their convention tests, the evaluation protocols and the training configurations are planned for public release. Model checkpoints are planned for public release upon acceptance.

\bibliography{references}
\bibliographystyle{iclr2027_conference}

\newpage
\appendix
\makeatletter\setlength{\@fptop}{0pt}\setlength{\@fpsep}{10pt plus 2pt}\setlength{\@fpbot}{0pt plus 1fil}\makeatother  %
\setcounter{topnumber}{4}\setcounter{totalnumber}{6}\renewcommand{\topfraction}{0.95}\renewcommand{\textfraction}{0.03}\renewcommand{\floatpagefraction}{0.8}
\section{Training runs and evaluation conventions}
\label{app:training}
The final model descends from the public MapAnything checkpoint through domain adaptation on first-generation renders (35 epochs), Stage 2 with online camera sampling (100 epochs), and two Stage-3 runs on second-generation scenes: a 15-epoch run on an earlier 189-scene batch of the same generator rendered at $2048\times1024$, and a run on the 469-scene training split, trained in segments of at most 125 epochs, of which the checkpoint after 859 epochs is used; the final weights interpolate the two with weights 0.25 and 0.75; both parents and the interpolation script will be released upon acceptance.
Tables~\ref{tab:realpano} and~\ref{tab:squeeze} also report the checkpoint after the first 90 epochs of the long run and an earlier interpolation with the same weights whose late parent is the checkpoint after 359 epochs.
Stage 2 gives each view of a tuple its own bucket, in a fixed pattern set by the batch bucket and $K$; Stage 3 and all evaluations use one bucket per tuple.
All stages use AdamW with weight decay 0.05 and cosine decay to 1\% of the peak learning rate. Stage 1 follows the official schedule, with peak rates of $10^{-4}$ ($5\times10^{-6}$ for the encoder) and 7 warm-up epochs; Stages 2 and 3 use $1.6\times10^{-5}$ ($8\times10^{-7}$ for the encoder) in bf16 autocast, with 15 warm-up epochs in Stage 2 and one in Stage 3. Each segment of the long run is a new run started from the weights of the previous segment, so the schedule restarts in every segment.
Reported results use isolated checkpoint evaluation and explicit preprocessing and panorama-routing settings; scripts and logs will be included in the release.

\section{\enginename{} in detail}
\label{app:engine}
This appendix gives the parameters behind Section~\ref{sec:engine} and the complete procedure for one training tuple (Algorithm~\ref{alg:delivery}).

\paragraph{Renders.}
First-generation panoramas are rendered at $2160\times1080$; the 37,010 second-generation poses at $3072\times1536$ with 128 samples per pixel.
In the first generation, narrow rectilinear views are re-rendered from the sharpest pinhole pack (18--40 pixels per degree) that covers the target and from the panorama (6 pixels per degree) otherwise; the second generation renders every view from its panorama (8.5 pixels per degree; 5.7 for the $2048\times1024$ batch of the 15-epoch run).

\paragraph{Camera models.}
Table~\ref{tab:cammodels} lists the seven models with their sampling weights and field-of-view ranges.
The weights give a rectilinear majority with a deliberate share of fisheye and panoramic views.
\begin{table}[h]
\centering
\caption{Camera models of the manifold with their sampling weights and initial field-of-view ranges. Rectilinear draws are log-normal around $80^\circ$ (log-space standard deviation 0.32) and clipped to the listed range; fisheye models and spherical crops are drawn log-uniformly. The parameter is diagonal for rectilinear models, nominal across the image circle for fisheye models (equidistant focal length, before distortion) and horizontal for spherical crops. Tuples in the aimed mode raise the lower bound to $90^\circ$, and the source-resolution floor and the covisibility repair can adjust a draw. Full panoramas span $360^\circ$ and receive a random $SO(3)$ content rotation.}
\label{tab:cammodels}
\footnotesize
\setlength{\tabcolsep}{6pt}
\begin{tabular}{@{}llcc@{}}
\toprule
Camera model & Family & Sampling weight & Field-of-view parameter \\
\midrule
OpenCV & Rectilinear with distortion & 0.40 & $48$--$120^\circ$ \\
Pinhole & Rectilinear & 0.15 & $48$--$95^\circ$ \\
Fisheye624 & Fisheye & 0.13 & $120$--$210^\circ$ \\
EUCM & Fisheye & 0.07 & $90$--$180^\circ$ \\
Mei & Fisheye & 0.05 & $140$--$200^\circ$ \\
Spherical crop & Spherical & 0.05 & $110$--$300^\circ$ \\
Full panorama & Spherical & 0.15 & $360^\circ$ \\
\bottomrule
\end{tabular}
\end{table}

\paragraph{Photographic priors.}
Principal-point shifts: 90\% Gaussian with $\sigma=1.2\%$ of the image size clipped at $\pm5\%$, of the order of the decentering reported for compact and phone cameras~\citep{sanzablanedo2010,patonis2023}, and 10\% uniform to $\pm22\%$ to cover zoom lenses and off-centre crops~\citep{clarke1998,perspectivefields}.
Roll: 90\% a zero-centred two-component Cauchy mixture (scales $0.06^\circ$ and $5.7^\circ$, weights 1/3 and 2/3) inspired by the roll model of~\citet{holdgeoffroy2023}, clipped at $\pm30^\circ$, 10\% uniform over $0$--$360^\circ$; tilt: 90\% Rayleigh with $\sigma=4^\circ$ clipped at $12^\circ$, our choice for a near-level indoor prior, 10\% uniform over $0$--$12^\circ$.
Optics layer: modulation-transfer randomisation on half of the views (downsampling factor log-uniform in $[0.35,1]$), lateral chromatic aberration, $\cos^4$ vignetting and sensor noise, a synthetic degradation pipeline in the manner of~\citet{realesrgan}; without it, sharpness would reveal the source asset and hence the field of view, a shortcut in the sense of~\citet{geirhos2020shortcut}.
The mutual information between field of view and sharpness, estimated on the first-generation source mix, falls from 0.557 to 0.251 bits.

\paragraph{Covisibility check and repair.}
The check renders the sampled views at 96 pixels on the long side with $2\times$ supersampling and recomputes $c_{ij}$ with Eq.~(\ref{eq:covis}).
A tuple whose graph is not connected at 0.25 is repaired by moving its disconnected views in three steps (0.5, 0.9 and 1.0 of the way in total) toward a wide field of view, a square aspect and no tilt or principal-point offset.
A tuple that still fails is rejected and the scene redrawn once; after that the sample falls back to the unaugmented renders, selected by connected walks on their precomputed covisibility (0.25), or by per-camera-pair thresholds when the walks fail (10.6\% of logged Stage-2 tuples, 14\% of views).
Kept views are rendered once more at their target resolution with $2\times$ supersampling.

\paragraph{Tuple-size schedule.}
The first epoch uses $K=4$ only, and a 15-epoch cosine ramp opens the distribution to its target over $K=2,\dots,8$ (0.08, 0.12, 0.15, 0.18, 0.18, 0.15, 0.14).
Each step holds $\lfloor 48/K\rfloor$ tuples per GPU on four GPUs (at most 48 images) and $\lceil\lfloor 48/K\rfloor/2\rceil$ on eight (at most 25 images).

\begin{algorithm}[h]
\caption{One training tuple: from panorama renders to the network input}
\label{alg:delivery}
\footnotesize
\begin{algorithmic}[1]
\Require scene with panorama renders $\{E_k\}$ (RGB, rays, depth, mask) and offline covisibility $c_{kl}$; tuple size $K$; bucket $(h,w)\in\mathcal{B}$
\Ensure input tensors $\{\tilde I_i,a_i,\pi_i\}_{i=1}^{K}$ and targets $\{\mathbf{r}_i,d_i,\mathbf{R}_i,\mathbf{t}_i,m_i\}_{i=1}^{K}$
\State draw the plan mode: aimed with probability 0.55 (converging on a common surface point with probability 0.90, parallel toward it otherwise; fields of view from $90^\circ$), native orientation otherwise
\State $(k_1,\dots,k_K)\gets\Call{ConnectedWalk}{c,K,\tau{=}0.40}$; in the aimed mode relax to $\tau{=}0.20$ if no chain of length $K$ exists; if no plan exists, retry in the other mode \Comment{distinct optical centres}
\For{$i=1$ \textbf{to} $K$}
  \State draw the camera model $\mu_i$, field of view $\theta_i$, distortion, principal point, roll and tilt from Table~\ref{tab:cammodels} and the priors above
  \State $\mathbf{R}^{\mathrm{c}}_i\gets$ gaze rotation or native orientation; \textbf{if} $\mu_i$ is a full panorama \textbf{then} $\mathbf{R}^{\mathrm{c}}_i\sim SO(3)$
  \State $V_i\gets\Call{Resample}{E_{k_i},\mu_i,\theta_i,\mathbf{R}^{\mathrm{c}}_i,\text{96 px}}$ \Comment{rays analytically; RGB, depth and mask by ray lookup}
\EndFor
\State $c_{ij}\gets\Call{Covis}{\{V_i\}}$ by Eq.~(\ref{eq:covis}); $\mathcal{G}\gets$ graph with edges $c_{ij}>0.25$
\For{$f\in(0.5,\,0.8,\,1.0)$ \textbf{while} $\mathcal{G}$ is not connected}
  \State pull each view outside the largest component a fraction $f$ of the remaining way toward the repair target (wide field of view, square aspect, no tilt, centred); re-render at 96 px; recompute $\mathcal{G}$
\EndFor
\If{$\mathcal{G}$ is not connected} \Return \textsc{reject} \Comment{redraw the scene once, then fall back to the native renders}
\EndIf
\For{$i=1$ \textbf{to} $K$}
  \State $V_i\gets\Call{Resample}{E_{k_i},\mu_i,\theta_i,\mathbf{R}^{\mathrm{c}}_i,\text{target resolution},2{\times}\text{ supersampling}}$
  \State $(H_i,W_i)\gets$ shape of $V_i$;\ $a_i\gets\log(W_i/H_i)$;\ $\pi_i\gets[\text{azimuth span of the middle row}>350^\circ]$;\ \textbf{if} $\pi_i$ \textbf{then} $a_i\gets\log 2$
  \State $\tilde I_i\gets$ RGB of $V_i$ resized to $(h,w)$; rays, depth and mask resampled through the same map (nearest for depth and mask)
\EndFor
\State \Return $\{\tilde I_i,a_i,\pi_i\}_{i=1}^{K}$ and the targets
\end{algorithmic}
\end{algorithm}
At inference only the last loop runs, on the real images, with the bucket chosen from the mean aspect ratio of the tuple and $\pi_i$ from the detector of Appendix~\ref{app:detector}.

\section{Panorama detector}
\label{app:detector}
The image is resized to 128 rows (width in proportion).
With $\mathrm{MAE}(\cdot,\cdot)$ the mean absolute colour difference between two columns and $\bar\sigma$ the mean horizontal colour standard deviation of a band of rows,
\begin{equation}
\rho_{\mathrm{seam}}=\frac{\mathrm{MAE}\big(I_{:,W},\,I_{:,1}\big)}{P_{90}\big\{\mathrm{MAE}(I_{:,k},I_{:,k+1})\big\}_{k<W}},\qquad
\rho_{\mathrm{pole}}=\frac{\tfrac12\big(\bar\sigma_{\mathrm{top}}+\bar\sigma_{\mathrm{bottom}}\big)}{\bar\sigma_{\mathrm{middle}}},
\label{eq:detector}
\end{equation}
over the top and bottom $H/16$ rows and the middle $H/8$ rows.
In a full panorama the last and first columns are adjacent in content, so $\rho_{\mathrm{seam}}$ is small, and the pole rows compress all longitudes into near-constant colour, so $\rho_{\mathrm{pole}}$ is small.
The guard $\eta$ is the fraction of rows in the central half whose 1\%-wide left and right edge strips both hold mostly non-black pixels; it rejects circular fisheyes, whose black corners would satisfy both ratios.
The flag is $\pi_i=[\eta\ge0.5]\wedge[\rho_{\mathrm{seam}}<3]\wedge[\rho_{\mathrm{pole}}<0.55]$.
Table~\ref{tab:detector} lists the error counts on the two headline sets and, in its caption, how the other sets are routed.

\begin{table}[h]
\centering
\caption{Full-panorama detector audit on the two headline sets with our detector (two image cues plus the black-circle guard, thresholds fixed on training renders): panoramas missed and other views flagged as panoramas. The 2D3DS panorama tuples are also routed with the detector; single panoramas and Matterport3D are flagged as panoramas throughout and the perspective video sets not at all. Counts on the other laser tracks, the 2D3DS panorama tuples and the single panoramas will accompany the code release.}
\label{tab:detector}
\footnotesize
\setlength{\tabcolsep}{3pt}
\begin{tabular}{@{}p{5.2cm}cp{2.6cm}p{3.2cm}@{}}
\toprule
Set & Views & Misses (panoramas not detected) & False alarms (other views flagged) \\
\midrule
Heterogeneous 2D3DS tuples, with guard & 503 & 0 / 189 & 2 / 314 \\
Laser mixed track, with guard & 96 & 0 / 24 & 0 / 72 \\
\bottomrule
\end{tabular}
\end{table}

\section{Additional tables and figures}
\label{app:tables}
\begin{figure}[h]
\centering
\includegraphics[width=0.98\textwidth]{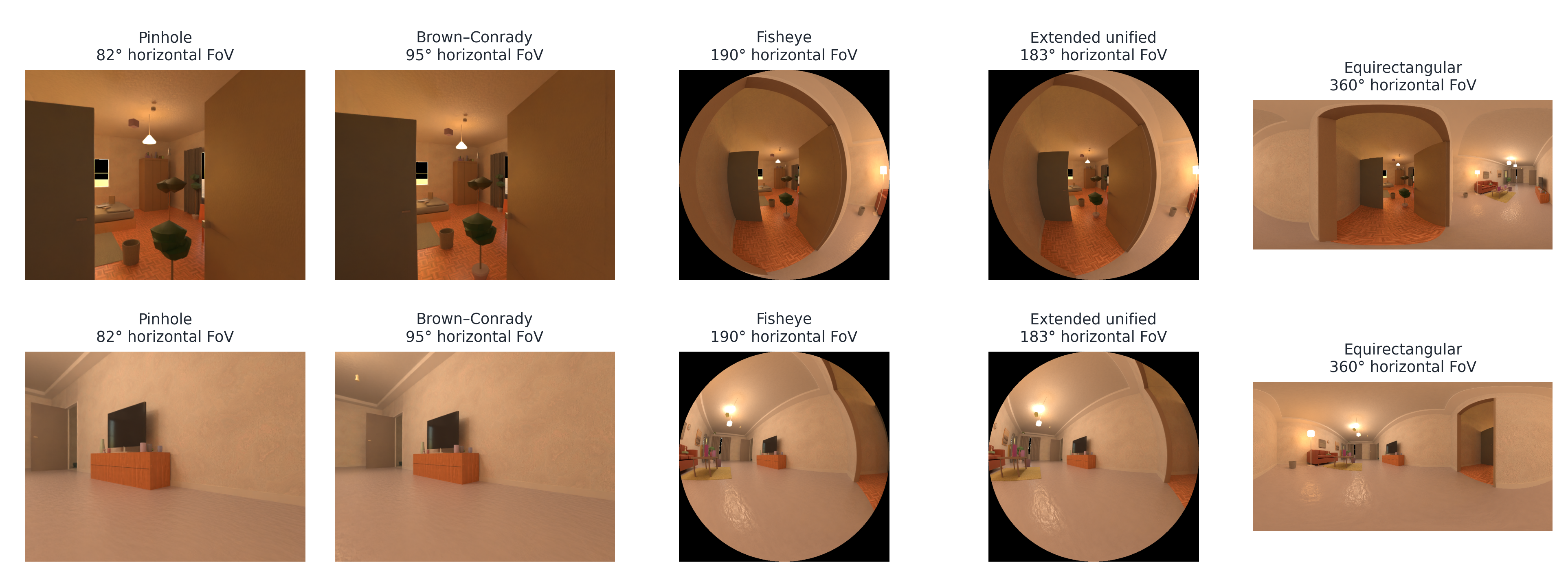}
\caption{Same-scene camera-model gallery from \enginename{}: two poses of one procedural scene (rows) rendered through five camera models (columns) at fixed illustrative fields of view; panel titles use the descriptive model names (Brown--Conrady for OpenCV, extended unified for EUCM).
The gallery illustrates the camera models; training tuples are assembled separately under the covisibility check of Section~\ref{sec:engine}.}
\label{fig:engine}
\end{figure}

\begin{figure}[h]
\centering
\includegraphics[width=\textwidth]{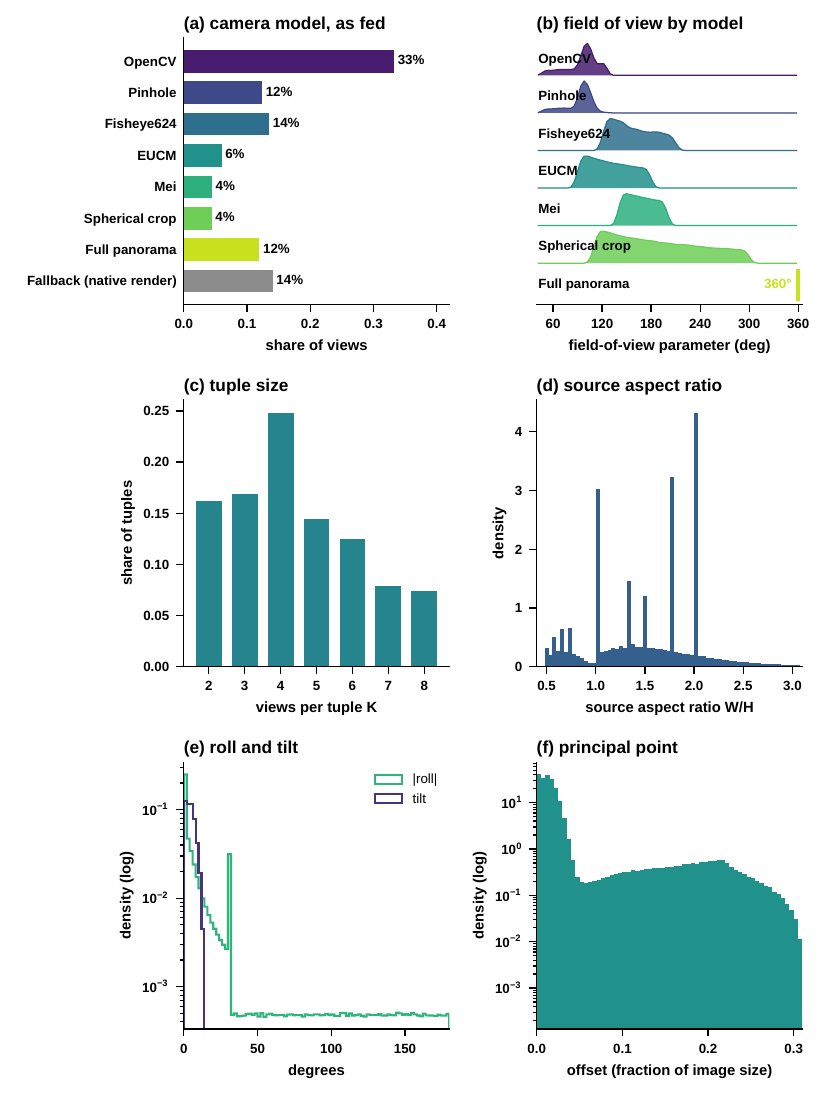}
\caption{As-fed distributions of the training stream, read from the per-tuple sampling log of one data-parallel rank of Stage 2 (438{,}001 tuples): (a) camera-model share, (b) the logged field-of-view parameter, one density strip per camera model (diagonal for rectilinear models, across the image circle for fisheye models, horizontal for spherical crops; full panoramas are $360^\circ$), (c) tuple size, (d) source aspect ratio, (e) roll magnitude and tilt, and (f) the principal-point offset magnitude; (e) and (f) use a log density so the 10\% uniform tails are visible. The fallback bar in (a) counts views delivered as unaugmented renders (Appendix~\ref{app:engine}).}
\label{fig:enginedist}
\end{figure}

\begin{figure}[h]
\centering
\includegraphics[width=0.86\textwidth]{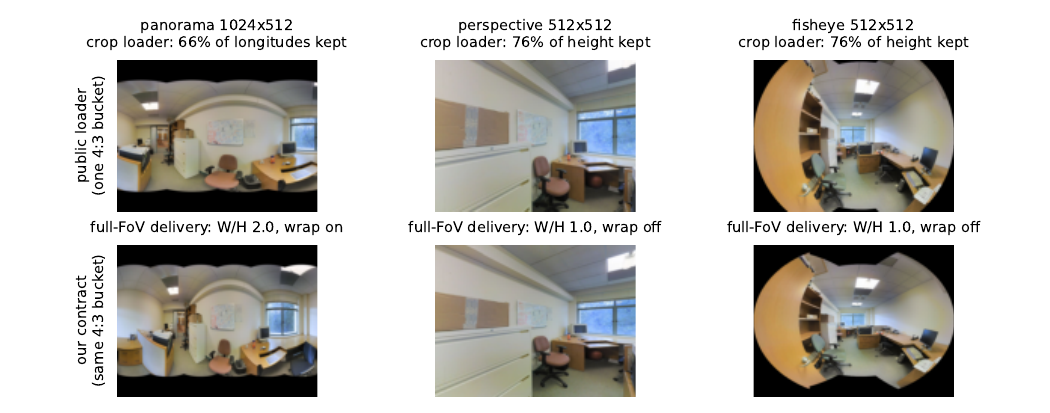}
\caption{What the network receives for one heterogeneous tuple.
The public loader picks one bucket from the mean aspect of the tuple, cover-scales and centre-crops every view, so the panorama loses a third of its longitudes and the square views lose a quarter of their height.
Our resizing maps every view anisotropically to the same bucket, keeps all content, and passes the aspect ratio and the panorama flag.}
\label{fig:contract}
\end{figure}

\begin{figure}[h]
\centering
\includegraphics[width=\textwidth]{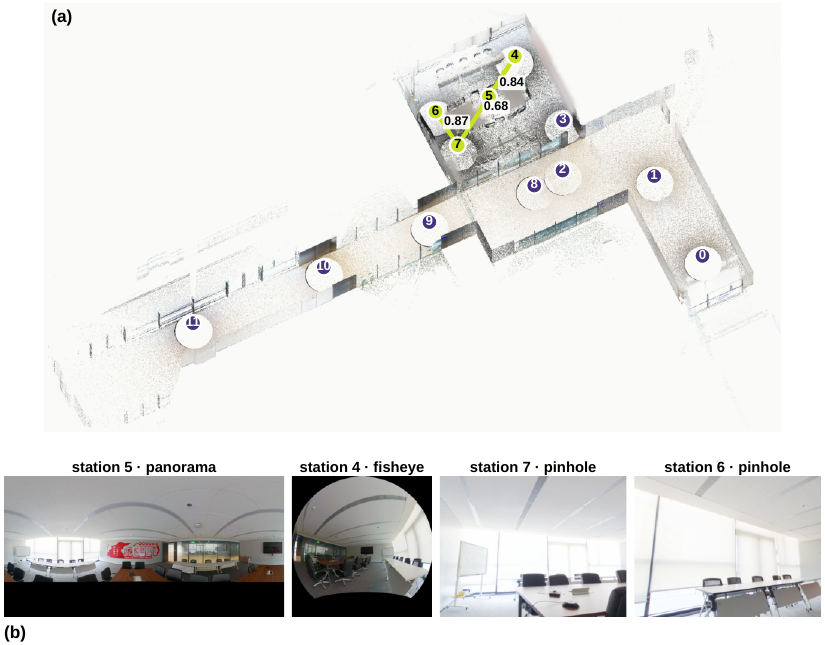}
\caption{The laser-scanned benchmark. (a) The merged point cloud of the twelve registered scanner stations, numbered 0--11: a corridor chain and a meeting room. The highlighted stations and links are one mixed tuple of the benchmark (stations 5, 4, 7 and 6); the numbers on the links are the covisibility between the station panoramas along the chain (0.84, 0.68 and 0.87). (b) The four views of that tuple as delivered to the models: a full panorama, a fisheye and two pinhole views resampled from the station scans.}
\label{fig:stations}
\end{figure}

Table~\ref{tab:datacmp} contrasts the training data of the compared methods with \enginename{}, Table~\ref{tab:controls} lists the resizing and routing controls of the final model, Table~\ref{tab:realpano} reports real panoramas, Table~\ref{tab:phaseb} the Stage-2 runs with and without the aspect-ratio embedding, Table~\ref{tab:squeeze} squeezed inputs of real panoramas, Table~\ref{tab:realpersp} the real perspective video sets, Table~\ref{tab:efficiency} the timing measurements, and Table~\ref{tab:t1views} the 2D3DS results by tuple size.
Figure~\ref{fig:engine} shows the camera models of \enginename{} on one scene, Figure~\ref{fig:enginedist} the as-fed distributions of the training stream.
Figure~\ref{fig:contract} shows the resizing on one tuple, and Figure~\ref{fig:stations} the laser-scanned benchmark's twelve stations with one of its tuples.

\begin{table}[t]
\centering
\caption{Training-data construction of the compared methods, from released code where available (X-Lens, a calibrated-rig depth method, omitted; test-time camera inputs are in Table~\ref{tab:contracts}). Of the compared methods, \enginename{} draws each view's camera from a continuous manifold, checks covisibility after the cameras are drawn, and uses no real image after the public perspective pretraining.}
\label{tab:datacmp}
\scriptsize
\setlength{\tabcolsep}{2pt}
\renewcommand{\arraystretch}{1.05}
\begin{tabular}{@{}p{1.55cm}p{2.45cm}p{2.35cm}p{2.4cm}p{1.8cm}p{2.65cm}@{}}
\toprule
 & Wid3R & CAM3R & Fisheye3R & PanoVGGT & \enginename{} (ours) \\
\midrule
Training data & 4 real + 5 synthetic sets & 4 real sets + views derived from panoramas & 3 real + 5 synthetic sets & 2 real + 2 synthetic panorama sets & Synthetic only, two procedural generations \\
Camera models seen & Pinhole, Fisheye624, Mei, equirectangular panorama (ERP); EUCM, Fisheye624, OpenCV splatted from pinhole (25\%) & Pinhole, fisheye (Aria; equidistant from panoramas), ERP & Pinhole; Kannala--Brandt fisheye distorted online from perspective frames; no 360$^\circ$ & ERP only & Continuous manifold: pinhole, OpenCV, Fisheye624, EUCM, Mei, spherical crops, full ERP \\
Camera identity within a tuple & One class per sample & Two-view pairs of mixed classes & Per-frame distortion, $p{=}0.5$ & N/A & Drawn per view, 2--8 views, distinct optical centres \\
Covisibility control & Position prior, no overlap test & Offline baseline and angle thresholds & GT-depth overlap $\geq25\%$ before distortion & Room or trajectory grouping & Checked after the cameras are drawn (mutual projection, depth agreement); failures repaired or dropped \\
\bottomrule
\end{tabular}
\end{table}

\begin{table}[t]
\centering
\caption{Resizing and routing controls for the final checkpoint and the public MapAnything weights on the two mixed-camera benchmarks; 2D3DS pose metrics per tuple, averaged over tuples. The Stage-1 row is the checkpoint adapted on first-generation renders with the public loader and loss, before the full-field-of-view (full-FoV) resizing, the embedding and the camera manifold. The detector reproduces the dataset annotation on every tuple of the mixed laser track (189 of 189 panoramas found on 2D3DS, 2 of 314 other views flagged). On the single-camera laser tracks the annotation gives 71.0 / 86.6 / 73.2 AUC@30 (panorama / pinhole / fisheye) against the detector's 71.0 / 85.7 / 73.2; on the 16 panorama tuples inside the covisibility envelope of Appendix~\ref{app:protocols} the final model reaches 92.6.}
\label{tab:controls}
\scriptsize
\setlength{\tabcolsep}{5pt}
\begin{tabular}{@{}llcc ccc@{}}
\toprule
 & & \multicolumn{2}{c}{2D3DS tuples} & \multicolumn{3}{c}{Laser mixed track} \\
\cmidrule(lr){3-4}\cmidrule(lr){5-7}
Method & Input resizing and panorama flag & mAA@30 & ATE $\downarrow$ & AUC@30 & Acc $\downarrow$ & Comp $\downarrow$ \\
\midrule
\sysname{} & Full-FoV resizing, detector flag & 80.4 & 0.62 & 79.4 & 0.142 & 0.367 \\
\sysname{} & Full-FoV resizing, annotation flag & 80.4 & 0.62 & 79.4 & 0.142 & 0.367 \\
\sysname{} & Public crop loader, annotation flag & 77.1 & 0.75 & 74.4 & 0.168 & 0.349 \\
Stage-1 checkpoint (renders only) & Public crop loader & 68.3 & 0.99 & 69.0 & 0.120 & 0.322 \\
MapAnything & Public crop loader & 16.4 & 1.48 & 24.5 & 0.255 & 0.830 \\
MapAnything & Full-FoV resizing, diagnostic & 15.4 & 1.45 & 19.7 & 0.272 & 0.808 \\
\bottomrule
\end{tabular}
\end{table}

\begin{table}[t]
\centering
\caption{Real panoramas. \emph{2D3DS tuples}: same-room tuples of 2--13 panoramas, 19 cases, seven areas; Wid3R reaches RRA@30 94.0 and RTA@30 96.1, PanoVGGT 100 and 100. \emph{Single panorama}: 40 panoramas of areas 5a and 5b, per-view Sim(3)-aligned Chamfer L1 in metres and median relative depth error. \emph{Matterport3D (MP3D) tuples}: 18 scans, 8 panoramas per tuple, pointmap accuracy and completeness in metres. $^\dagger$PanoVGGT trains on 2D3DS and Matterport3D, so its 2D3DS numbers are in-domain; our model, Wid3R and the open baselines are zero-shot here. Wid3R's published 79.9 on 2D3DS uses a different protocol and is not the number in this column.}
\label{tab:realpano}
\scriptsize
\setlength{\tabcolsep}{3pt}
\begin{tabular}{@{}p{2.7cm}p{1.4cm}cccc@{}}
\toprule
 & & 2D3DS tuples & \multicolumn{2}{c}{2D3DS single panorama} & MP3D tuples \\
\cmidrule(lr){3-3}\cmidrule(lr){4-5}\cmidrule(lr){6-6}
Model & Checkpoint & AUC@30 & Chamfer L1 $\downarrow$ & Depth rel.\ $\downarrow$ & Acc / Comp $\downarrow$ \\
\midrule
\sysname{} & Final & 75.4 & 0.296 & 0.195 & 0.237 / 0.962 \\
\sysname{} & 90-epoch run & 70.5 & 0.328 & 0.315 & 0.249 / \textbf{0.952} \\
Wid3R (camera type given) \citep{wid3r} & Released & 92.2 & 0.090 & 0.044 & -- \\
PanoVGGT$^\dagger$ \citep{panovggt} & Released & \textbf{99.9} & \textbf{0.063} & \textbf{0.028} & -- \\
MapAnything \citep{mapanything} & Public & 7.0 & 0.852 & 0.520 & \textbf{0.211} / 2.096 \\
$\pi^3$ \citep{pi3} & Public & 8.4 & -- & -- & -- \\
VGGT \citep{vggt} & Public & 5.3 & -- & -- & -- \\
DUSt3R \citep{dust3r} & Public & 1.2 & -- & -- & -- \\
MASt3R \citep{mast3r} & Public & 2.2 & -- & -- & -- \\
\bottomrule
\end{tabular}
\end{table}

\begin{table}[t]
\centering
\caption{Stage-2 runs with and without the aspect-ratio embedding. Both start from the Stage-1 checkpoint and train the Stage-2 configuration (first-generation scenes, full-field-of-view resizing, online camera sampling, 100 epochs, seed 0); the only configuration difference is the embedding. Same protocol as the main tables; 2D3DS pose metrics per tuple.}
\label{tab:phaseb}
\scriptsize
\setlength{\tabcolsep}{3pt}
\begin{tabular}{@{}lccccc@{}}
\toprule
 & \multicolumn{3}{c}{Heterogeneous 2D3DS tuples} & \multicolumn{2}{c}{Laser tracks, AUC@30} \\
\cmidrule(lr){2-4}\cmidrule(lr){5-6}
Stage-2 run & RRA@30 & RTA@30 & mAA@30 & Mixed & Panorama \\
\midrule
With the embedding & 91.9 & 93.3 & 63.8 & 65.4 & 45.0 \\
Without the embedding & 90.0 & 92.8 & 60.6 & 61.7 & 46.7 \\
\midrule
\multicolumn{6}{@{}l}{2D3DS panorama tuples, AUC@30: native 58.1 against 53.9; squeezed to 16:9, 63.8 against 53.9} \\
\bottomrule
\end{tabular}
\end{table}

\begin{table}[t]
\centering
\caption{Squeezed input of real panoramas (2D3DS panorama tuples, AUC@30). The rows squeeze every panorama to the stated shape before it enters the pipeline. The panorama detector supplies the content aspect and panorama wrap in the final-model rows. At fixed 16:9 pixels, withholding the aspect-ratio embedding lowers the final model from 78.6 to 42.4. Dashes mark shapes that were not run for that checkpoint.}
\label{tab:squeeze}
\small
\setlength{\tabcolsep}{5pt}
\begin{tabular}{@{}llccc@{}}
\toprule
Input shape & Aspect input & 90-epoch run & Earlier interpolation & Final \\
\midrule
Native 2:1 & Content aspect (2.0) & 70.5 & 74.9 & 75.4 \\
16:9 squeeze & Tensor aspect (1.78) & 66.4 & 68.8 & -- \\
16:9 squeeze & Content aspect, from the detector (2.0) & 75.9 & 78.8 & 78.6 \\
16:9 squeeze & Embedding withheld & -- & -- & 42.4 \\
4:3 squeeze & Content aspect, from the detector (2.0) & 71.8 & -- & 76.2 \\
1:1 squeeze & Content aspect, from the detector (2.0) & 67.9 & 75.1 & 76.9 \\
\bottomrule
\end{tabular}
\end{table}

\begin{table}[t]
\centering
\caption{Real perspective video with centimetre baselines, outside the room-scale tuples of the engine (AUC@30, same tuples, ground truth and scorer for every model). Rotation is solved by every model (RRA@30 of 99--100); the gap is translation direction under micro-baselines. Our row uses the final checkpoint with no view flagged as a panorama. VGGT trains on Replica and ADT.}
\label{tab:realpersp}
\small
\setlength{\tabcolsep}{5pt}
\begin{tabular}{@{}lcccccc@{}}
\toprule
Set (tuples) & \sysname{} & MapAnything & VGGT & $\pi^3$ & DUSt3R & MASt3R \\
\midrule
Replica (48) & 69.0 & 78.4 & 93.5 & \textbf{98.5} & 87.1 & 98.0 \\
ADT pinhole (80) & 39.4 & 60.8 & 53.7 & 80.4 & 49.3 & \textbf{83.4} \\
ADT fisheye (80) & 38.4 & 41.0 & 42.2 & \textbf{46.1} & 28.4 & 34.0 \\
\bottomrule
\end{tabular}
\end{table}

\begin{table}[t]
\centering
\caption{Efficiency. Top: full pipelines per tuple on one RTX 5090, including image loading, first tuple excluded as warm-up. Bottom: the network forward pass alone at 518 px (random inputs pre-built on the GPU, 30 iterations after 8 warm-up iterations, CUDA-event timing). A six-point log-log fit gives exponents 1.09 (bf16) and 1.28 (fp32); the local slope rises with $N$, so this is an empirical scaling and not a complexity statement.}
\label{tab:efficiency}
\footnotesize
\setlength{\tabcolsep}{3pt}
\begin{tabular}{@{}p{4.2cm}cccccc@{}}
\toprule
Per-tuple wall clock, I/O included (s) & \sysname{} & MapAnything & DUSt3R & MASt3R & & \\
\midrule
Replica, 8 views (RTX 5090) & 0.66 & 0.66 & 5.54 & 8.36 & & \\
Laser mixed track, 4 views (RTX 5090) & 0.48 & -- & 3.65 & 4.83 & & \\
\midrule
Network forward pass, $N$ views (RTX 5090) & 2 & 4 & 8 & 16 & 24 & 32 \\
\midrule
BF16 mean (ms) & 67.5 & 126.9 & 266.8 & 583.0 & 965.8 & 1359.9 \\
FP32 mean (ms) & 129.2 & 267.7 & 610.9 & 1558.4 & 2867.0 & 4479.4 \\
Peak memory BF16 (GB) & 7.85 & 9.03 & 11.42 & 16.20 & 20.98 & 25.77 \\
\bottomrule
\end{tabular}
\end{table}

\begin{table}[h]
\centering
\caption{Heterogeneous 2D3DS tuples by tuple size; the final model leads on 79 of the 88 tuples. Per-tuple mAA@30 and per-tuple accuracy at $15^\circ$, averaged over the tuples of each size range, for the final model and for Wid3R given the camera type of every view; the last column counts the tuples on which our per-tuple mAA@30 is higher. Training tuples have at most eight views.}
\label{tab:t1views}
\scriptsize
\setlength{\tabcolsep}{4pt}
\begin{tabular}{@{}lr ccc ccc c@{}}
\toprule
 & & \multicolumn{3}{c}{\sysname{} (pixels only)} & \multicolumn{3}{c}{Wid3R (camera type given)} & \\
\cmidrule(lr){3-5}\cmidrule(lr){6-8}
Views & Tuples & mAA@30 & RRA@15 & RTA@15 & mAA@30 & RRA@15 & RTA@15 & Ours ahead \\
\midrule
3 & 36 & \textbf{87.5} & \textbf{99.1} & \textbf{100.0} & 53.6 & 91.7 & 64.8 & 35/36 \\
4 & 13 & \textbf{85.0} & \textbf{98.7} & \textbf{100.0} & 56.3 & 87.2 & 72.4 & 13/13 \\
5--8 & 28 & \textbf{79.7} & \textbf{92.9} & \textbf{93.0} & 54.0 & 88.8 & 67.3 & 25/28 \\
9--14 & 7 & \textbf{67.9} & \textbf{83.9} & \textbf{81.2} & 56.9 & 83.6 & 70.1 & 6/7 \\
15--24 & 4 & 29.3 & 46.7 & 49.2 & \textbf{52.1} & \textbf{71.1} & \textbf{71.2} & 0/4 \\
\bottomrule
\end{tabular}
\end{table}

\clearpage

\section{Downstream preview: cross-camera correspondences}
\label{app:corr}
\begin{figure}[h]
\centering
\includegraphics[width=0.84\textwidth]{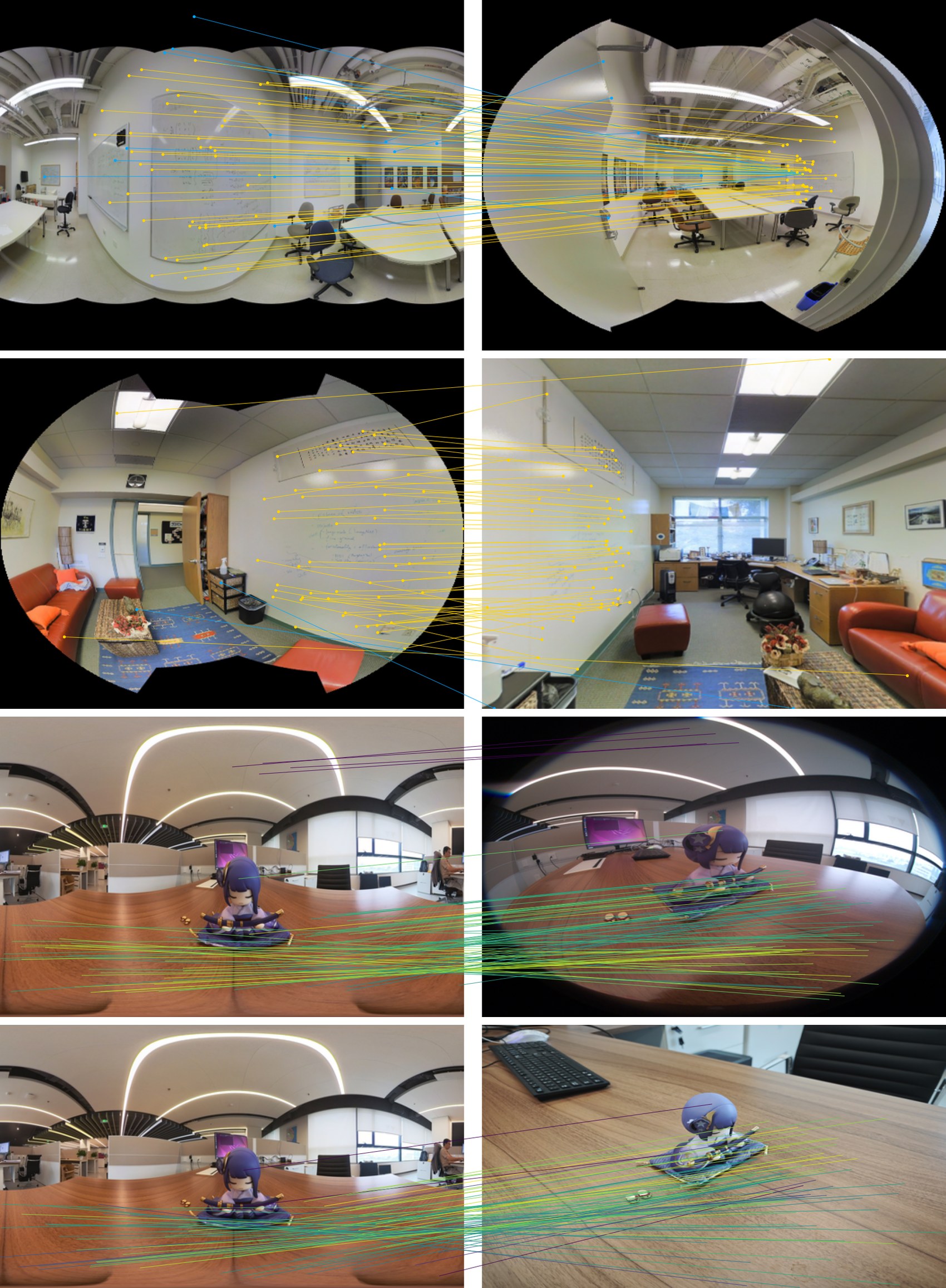}
\caption{Cross-camera correspondences of the final model, obtained as mutual three-dimensional nearest neighbours between predicted world points. Rows one and two: real 2D3DS pairs with ground truth, panorama to fisheye and fisheye to perspective; yellow lines end within $3^\circ$ of the true target, blue lines do not. Rows three and four: pairs without ground truth from a consumer panorama, a fisheye and a phone photograph of the same desk, coloured by the predicted three-dimensional agreement of the mutual match (brighter is closer).}
\label{fig:corr}
\end{figure}

A model that maps the same physical point to the same world coordinate regardless of the camera that observed it yields correspondences without a matcher: for a pixel of view $A$, its match in view $B$ is the pixel whose predicted world point is the three-dimensional nearest neighbour of $A$'s predicted point, and a match is kept when it is mutual.
No matching head is trained and no descriptor is compared.
Figure~\ref{fig:corr} shows such matches from the final model on real 2D3DS pairs across camera kinds and on pairs of a consumer panorama, a fisheye and a phone photograph of the same desk.

\section{Protocol details}
\label{app:protocols}
Pointmaps use accuracy, completeness and normal consistency after one alignment per tuple (the alignment of~\citet{umeyama}, a least-squares scale and shift, then ICP~\citep{besl1992icp}).
On the two mixed-camera benchmarks our model receives every view of a tuple at the common bucket by full-field-of-view resizing, with the aspect-ratio embedding and the image-detected panorama wrap; on the 2D3DS panorama, Matterport3D, Replica and ADT sets it uses the public crop loader.
The open baselines (MapAnything, VGGT, $\pi^3$, DUSt3R, MASt3R) run with their official loaders and checkpoints on the same tuples, ground truth and scorer.
Wid3R runs with its released weights and its official settings, which give the network the camera type of every view; its camera-frame convention was fixed once on a calibration split disjoint from the heterogeneous benchmark (2D3DS areas 1--4) and then frozen.
\paragraph{Pose metrics.}
Pose metrics are computed over ordered view pairs: RRA@30 and RTA@30 are the fractions of pairs whose relative rotation error and translation-direction error (sign-agnostic,~\citealp{posediffusion}) are below 30$^\circ$; mAA@30 averages, over thresholds from 1$^\circ$ to 30$^\circ$, the fraction of pairs with both errors below the threshold, and AUC@30 integrates the same curve; ATE is the trajectory error after one Sim(3) alignment per tuple.
On the 2D3DS benchmark the metrics are computed per tuple and averaged over tuples.
Pooled over all 4{,}044 ordered pairs instead, they weight each tuple by its number of pairs, and the four tuples of 15--24 views hold 49\% of the pairs (Table~\ref{tab:t1views}); pooled mAA@30 / RRA@30 / RTA@30 are 54.2 / 74.5 / 81.0 for \sysname{}, 54.7 / 88.3 / 84.2 for Wid3R, 18.9 / 41.9 / 58.7 for $\pi^3$, 14.4 / 47.0 / 49.8 for MapAnything and 9.0 / 32.5 / 44.1 for VGGT.
Relative pose metrics need no alignment; ATE and the pointmap metrics are computed after one alignment per tuple.
The predicted metric scale is measured without alignment as the ratio of ground-truth to predicted camera-centre distances, median over the view pairs of a tuple and then over tuples: 1.14 on 2D3DS (74\% of tuples within $\pm$20\%, 91\% within $\pm$30\%) and 1.26--1.34 on the four laser tracks for the final model, against 0.63 and 0.70 for the public MapAnything weights on 2D3DS and the laser mixed track.
\paragraph{2D3DS panoramas.}
The panorama tuples of Table~\ref{tab:realpano} apply the sampling that Wid3R describes for 2D3DS (10--30 images per scene, ten draws per scene) to rooms: a random subset of 10--30 panoramas per room (all of them in smaller rooms), ten draws per room over the seven areas, shuffled with seed 0 and truncated to 20; one small room is drawn twice with the same two panoramas, and the repeat is counted once.
For single panoramas, the loader scales each panorama to $518\times259$ and crops seven rows, and the ground truth is cropped the same way.
\paragraph{Laser-scanned benchmark.}
The scene is a multi-room office (a corridor chain and a meeting room) scanned from twelve stations with a Leica BLK360 G2 and registered in Cyclone; each station provides a $1536\times3072$ panorama with along-ray laser depth in the registered frame.
The laser benchmark gates consecutive station pairs on the covisibility of the benchmark views themselves, estimated from 1,500 sampled points per direction and symmetrised by the minimum of the two directions (depth agreement within the larger of 10\,cm and 3\%): 0.25 for the pinhole and fisheye tracks, 0.10 for the panorama track and 0.15 for mixed tuples.
\paragraph{Covisibility envelope.}\label{sec:envelope}
Registration failures concentrate at low pairwise covisibility. On the laser panorama track, the eight tuples on which the final model misses a view pair at $30^\circ$ (in rotation or translation) have a median minimum pairwise covisibility of 0.042 between their station panoramas (mutual projection of the scans, 10\,cm agreement), the sixteen others 0.424, and five of the eight lie below 0.05; camera-sampled training tuples are checked for connectivity at 0.25 (Section~\ref{sec:engine}). The envelope subsets of Table~\ref{tab:controls} add a floor of 0.10 on the benchmark covisibility of every view pair, with consecutive gates of 0.25 on the panorama track (16 tuples) and 0.20 on the mixed track (13 tuples).
\paragraph{Routing and controls.}
Headline numbers infer the binary full-panorama switch from the pixels; controls that instead use the dataset's panorama annotation are marked explicitly.
Our self-run VGGT and $\pi^3$ rotation accuracies, pooled over pairs as in CAM3R's public evaluation code and paper, match the values it publishes for them (32.5 against 31.8 and 41.9 against 40.0).
\paragraph{Regenerating the data.} The code release will include the commands that regenerate a first-generation scene (eight poses, five pinhole, two fisheye and one panoramic render at 64 samples per pixel) and a second-generation scene (panoramas at $3072\times1536$ and 128 samples per pixel), build shards and offline covisibility, draw one covisibility-checked tuple and run ten optimiser steps as a pipeline check; the second-generation split (469/15/15, seed 20260729) is reproduced exactly by a script planned for release, and the optional 935-file CC0 and public-domain asset bundle will include a per-file provenance inventory.

\end{document}